\documentclass[letterpaper, 10 pt, conference]{ieeeconf}  

\IEEEoverridecommandlockouts                              

\usepackage{graphics} 
\usepackage{epsfig} 
\usepackage{mathptmx} 
\usepackage{times} 
\usepackage{amsmath} 
\usepackage{amssymb}  
\usepackage{todonotes}
\usepackage{etoolbox}
\usepackage{algorithm}
\usepackage{algpseudocode}
\usepackage{subcaption}

\usepackage[font=small,labelfont=bf]{caption} 

\usepackage{hyperref}

\usepackage{sidecap}
\usepackage{array}

\title{\LARGE \bf
Efficient Incremental Penetration Depth Estimation \\ between Convex Geometries
}

\author{Wei Gao
\thanks{Mech-Mind Robotics. Email: gaowei19951004@hotmail.com
}%
}

\begin{document}

\maketitle
\thispagestyle{empty}
\pagestyle{empty}

\begin{abstract}

Penetration depth (PD) is essential for robotics due to its extensive applications in dynamic simulation, motion planning, haptic rendering, etc. The Expanding Polytope Algorithm (EPA) is the de facto standard for this problem, which estimates PD by expanding an inner polyhedral approximation of an implicit set. In this paper, we propose a novel optimization-based algorithm that incrementally estimates minimum penetration depth and its direction. One major advantage of our method is the capability to be warm-started by leveraging the spatial and temporal coherence. 
This coherence emerges naturally in many robotic applications (e.g., the temporal coherence between adjacent simulation time knots). As a result, our algorithm achieves substantial speedup –- we demonstrate it is 5-30x faster than EPA on several benchmarks. Moreover, our approach is built upon the same implicit geometry representation as EPA, which enables easy integration into existing software stacks.
The code and supplemental document are available on: \href{https://github.com/weigao95/mind-fcl}{\textcolor{blue}{\underline{https://github.com/weigao95/mind-fcl}}}.

\end{abstract}

\section{Introduction}
\label{sec:intro}

Penetration depth (PD) is a distance measure that characterizes how much two overlapping shapes penetrate into each other.
PD is of significant importance to various robotic applications. For instance, 1) in dynamic simulations, PD is used to calculate the contact force~\cite{trinkle1997dynamic, mirtich1998rigid, todorov2014convex} in almost all rigid body contact models; 2) in motion planning, many planners~\cite{toussaint2018differentiable, schulman2013finding, kappler2018real} are designed to regulate (minimize) the PD to alleviate and avoid collisions; and 3) in haptic rendering~\cite{gregory2000six, mcneely2005six, kumar2015mujocohap}, PD is used to resolve the interactions between objects.
PD estimation is usually coupled with binary collision detection, which provides overlapping shape pairs as its input~\cite{coumans2016pybullet, pan2012fcl}. A common strategy to accelerate collision detection and PD estimation is to split the computation into two phases~\cite{ericson2004real}. The first is the ``broad phase'' which eliminates shape pairs that are too far away. The second ``narrow phase'' checks if the shape pairs passed the broad phase are really colliding, and computes the PD for the colliding pairs. This paper focuses on PD estimation in the narrow phase, as detailed below.

\begin{figure}[t]
\centering
\includegraphics[width=0.45\textwidth]{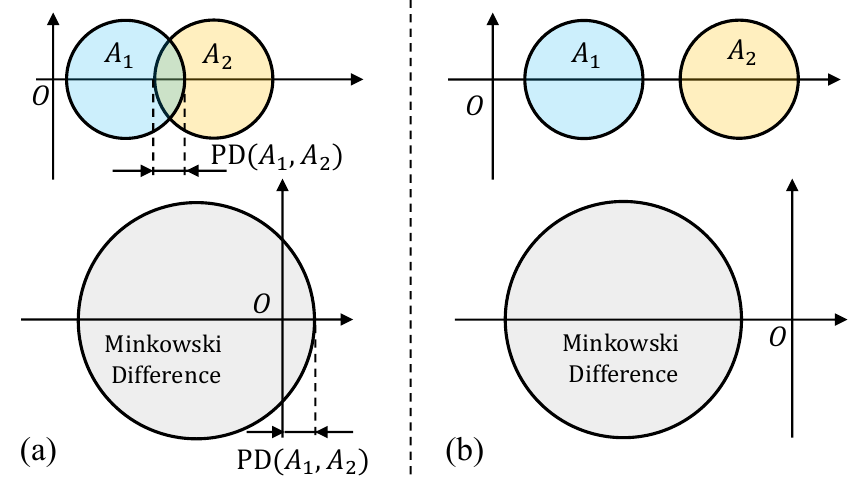}
\caption{\label{fig:penetration_illustration} An illustration of penetration depth and Minkowski Difference. (a) For two colliding shapes, their Minkowski Difference contains the origin $O$. The penetration depth PD$(A_1, A_2)$ is also the minimum distance from the origin $O$ to the boundary of their Minkowski Difference. (b) For two non-overlapping shapes, the origin $O$ is outside their Minkowski Difference and PD is undefined. }
\end{figure}

\vspace{1mm}
\noindent \textbf{Problem Formulation. } We investigate two closed convex shapes $A_1, A_2 \subset$ $R^n$. Usually, the space dimension $n=2$ or $n=3$, corresponds to 2-dimensional (2D) or 3-dimensional (3D) setups. Non-convex shapes can be handled by computing their convex-hull~\cite{melkman1987line} or performing convex decomposition~\cite{mamou2009simple}. Two shapes $A_1, A_2$ collide with each other if $A_1 \cap A_2 \neq \emptyset$. For two shapes $A_1, A_2$ that are in collision, the penetration depth $\text{PD}(A_1, A_2)$ is defined as
\begin{align}
\begin{split}
\label{equ:formulation}
    \text{PD}(A_1, A_2) = ~& \text{min}_{~d_{1, 2} \in R^n}~~~||d_{1, 2}||_2 \\
     &\text{subject to:}~~\text{interior}(A_1+d_{1, 2}) \cap A_2 = \emptyset
\end{split}
\end{align}
\noindent where $\| \cdot \|_2$ is L2-norm; $A_1 + d_{1,2}$ is the set obtained by applying a translation $d_{1, 2} \in R^n$ to each point in $A_1$. The displacement $d_{1, 2}^{*} \in R^n$ that is an optimal solution to Problem~(\ref{equ:formulation}) is denoted as \textit{minimum penetration displacement}. The unit vector $n_{1, 2}^{*} = \text{normalized}(d_{1, 2}^{*})$ is usually denoted as \textit{minimum penetration direction}. Obviously, $d_{1, 2}^{*} = n_{1, 2}^{*} \times \text{PD}(A_1, A_2)$. An illustration is shown in Fig~\ref{fig:penetration_illustration}.

\vspace{1mm}
\noindent \textbf{Related Works. } The most widely used method for PD estimation is the EPA~\cite{van2001epa}, which is coupled with the Gilbert-Johnson-Keerthi (GJK) algorithm~\cite{gilbert1988gjk} for collision detection. Both GJK and EPA operate on \textit{Minkowski Difference}~\cite{gilbert1988gjk}, which is a convex set constructed from original shapes $A_1$ and $A_2$ (a detailed explanation of Minkowski Difference is in Sec.~\ref{sec:preliminary}). Intuitively, EPA algorithm estimates the PD by building an inner polyhedral approximation of the Minkowski Difference. EPA algorithm requires a tetrahedron inscribed to the Minkowski Difference as the input, which is produced by GJK algorithm for each colliding shape pair. Built upon the Minkowski Difference formulation, GJK and EPA can handle arbitrary convex shapes such as convex polyhedra and basic primitives (\textit{i.e.,} spheres, boxes etc.). These desirable properties make EPA the de facto standard for PD estimation. However, an accurate PD estimation using EPA may require an inner polyhedral approximation with a lot of vertices and faces. In this paper, we take a different approach that incrementally estimates PD by solving an optimization problem. A key advantage of our method is to leverage spatial and temporal coherence in many robotic applications to warm-start the optimization. For instance, in dynamic simulation the object pairs tend to have very similar penetration direction and depth in consecutive time steps. Initializing our incremental PD algorithm by the information of the previous time knot can lead to substantial speed-up.

The idea of incrementally estimating the PD has also been explored in DEEP~\cite{kim2004incremental},  which seeks a locally optimal solution of PD by walking on vertices of Minkowski Difference. 
Compared with~\cite{kim2004incremental}, our method can handle arbitrary convex shapes, while \cite{kim2004incremental}~is restricted to convex polyhedra. Specifically, the proposed algorithm is capable of penetration computation between two primitive shapes or a primitive-polyhedra pair, while \cite{kim2004incremental} does not support it.
Moreover, experimental results in Sec.~\ref{sec:results} demonstrate the proposed method is about 2x faster than~\cite{kim2004incremental}.
In the broader context of penetration computation between non-convex geometries, several contributions~\cite{je2012polydepth, tang2014interactive, nawratil2009generalized} proposed to iteratively compute the tangent space of Minkowski Difference and project onto it, which is conceptually related to our method. Compared with them, the key technical distinction of our method is: 1) a novel method to construct the tangent space of Minkowski Difference for convex geometries (Sec.~\ref{subsec:alg_2}); and 2) an early termination mechanism to reduce the iterations and improve the performance (Sec.~\ref{subsec:alg_3}). 
%
Non-convex penetration algorithms~\cite{je2012polydepth, tang2014interactive, nawratil2009generalized} and convex ones~\cite{van2001epa, kim2004incremental} have different trade-off on problem complexity and computational performance. These two types of algorithms complement each other in practical applications.
Researchers have also proposed to explicitly construct the Minkowski Difference~\cite{dobkin1993computing} for PD estimation. However, the construction procedures tend to be computationally expensive.

\vspace{1mm}
\noindent \textbf{Contributions. }
Built upon the seminal works~\cite{gilbert1988gjk, van2001epa, jacobs2008mpr}, this paper proposes a novel incremental PD estimation algorithm between general convex shapes. In particular,

\begin{itemize}
    \item We formulate PD estimation as a ``Difference-of-Convex'' problem, which is usually solved by Sequential Quadratic Programming (SQP)~\cite{boyd2004convex}. The major challenge of applying SQP to PD estimation is the implicit geometry representation (the Minkowski Difference and \textit{Support Function} explained in Sec.~\ref{subsec:supportpre}). To address it, we propose a novel instantiation of SQP that utilizes a modified Minkowski Portal Refinement (MPR) algorithm~\cite{jacobs2008mpr} as a subroutine.
    \item We propose a novel shortcut mechanism that reduces the computation of the vanilla SQP procedure for PD estimation. Furthermore, we demonstrate that the algorithm can still converge to locally optimal solutions despite the shortcut mechanism.
    \item We experimentally evaluate our method on several benchmarks. Our method demonstrates a 5-30x speedup over EPA at a comparable accuracy.
    \item Our method can be easily integrated into existing software stacks and we provide an open-source implementation to facilitate its usage in robotic applications.
\end{itemize}

This paper is organized as follows: in Sec.~\ref{sec:preliminary} presents the preliminaries. Sec.~\ref{sec:algorithm} explains the PD algorithm. Sec.~\ref{sec:results} shows the results. Sec.~\ref{sec:conclusion} concludes.

\section{Preliminary}
\label{sec:preliminary}

\subsection{Minkowski Difference and Penetration Depth}
\label{subsec:mkdiffpre}

As mentioned in Sec.~\ref{sec:intro}, we investigate two closed, convex shapes $A_1, A_2 \subset$ $R^n$. Usually the dimension $n=2$ or $n=3$. The Minkowski Difference $\mathcal{D}_{1, 2}$ for $A_1$ and $A_2$ is defined as
\begin{align}
\begin{split}
\label{equ:mkdiff}
    \mathcal{D}_{1, 2} = A_1 - A_2 = \{v = v_1 - v_2| v_1 \in A_1, v_2 \in A_2 \}
\end{split}
\end{align}
%
The following properties hold for $\mathcal{D}_{1, 2}$ as proved in~\cite{van2001epa}:

\begin{itemize}
    \item $\mathcal{D}_{1, 2}$ is a closed convex set. 
    \item $A_1$ and $A_2$ overlap if the origin $O \in \mathcal{D}_{1, 2}$.
    \item For two overlapping shapes $A_1$ and $A_2$, their penetration depth $\text{PD}(A_1, A_2)$ defined in Problem~(\ref{equ:formulation}) corresponds to the minimum distance from $O$ to the \textit{boundary} of $\mathcal{D}_{1, 2}$.
\end{itemize}

\noindent An illustration is provided in Fig.~\ref{fig:penetration_illustration}. Using these properties, we can reformulate Problem~(\ref{equ:formulation}) in Sec.~\ref{sec:intro} into:
\begin{align}
\begin{split}
\label{equ:mkdiffopt}
    \text{PD}(A_1, A_2) = ~& \text{min}_{~v \in R^n}~~~||v||_2 \\
     &\text{subject to:}~~v \in \text{boundary}(\mathcal{D}_{1, 2})
\end{split}
\end{align}
\noindent Let the point $v^{*}$ on the boundary of $\mathcal{D}_{1, 2}$ be one optimal solution to Problem~(\ref{equ:mkdiffopt}). The minimum penetration displacement $d_{1, 2}^{*}$ in Sec.~\ref{sec:intro} can be computed as $d_{1, 2}^{*} = - v^{*}$ (in other words $\text{interior}(A_1 - v^{*}) \cap A_2 = \emptyset$). The minimum penetration direction in Sec.~\ref{sec:intro} can be computed as $n_{1, 2}^{*} = \text{normalized}(-v^{*})$.

\subsection{Support Function for Convex Shapes}
\label{subsec:supportpre}


\begin{figure}[t]
\centering
\includegraphics[width=0.38\textwidth]{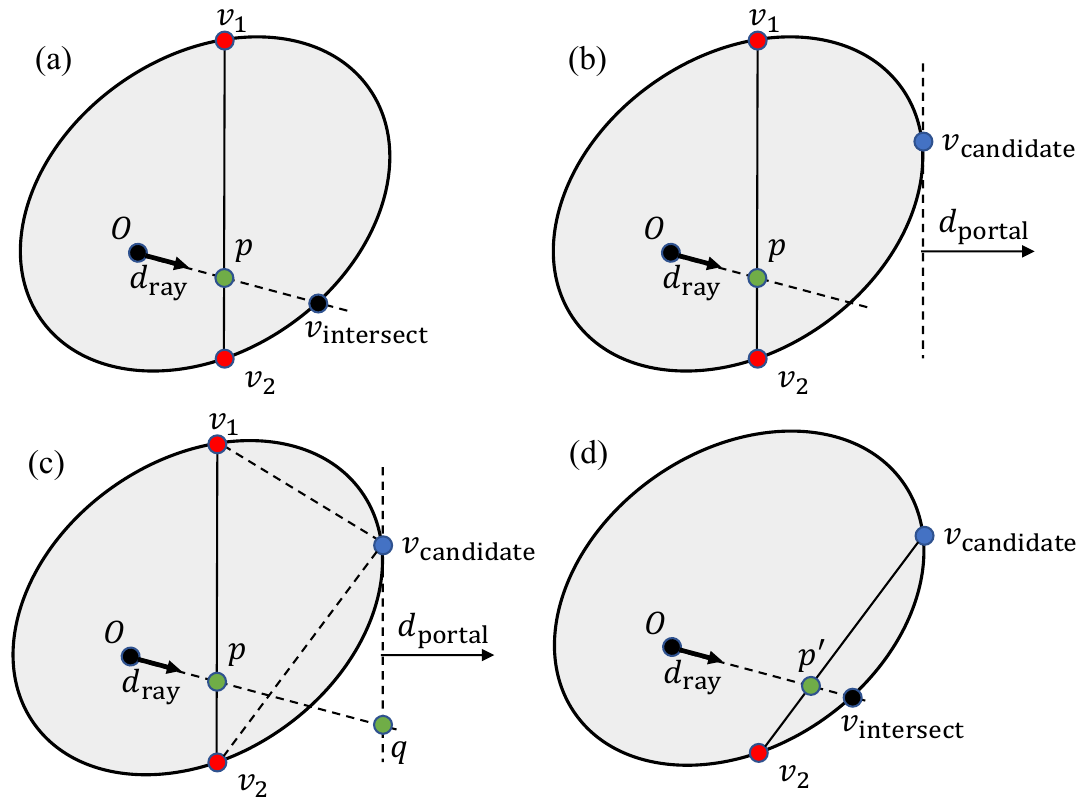}
\caption{\label{fig:mpr_illustration} An illustration of a modified MPR algorithm to compute the intersection point (the $v_{\text{intersect}}$ in (a)) of an origin ray with the boundary of the Minkowski Difference. The algorithm takes the ray direction $d_{\text{ray}}$ as the input. In each iteration, the algorithm maintains and updates a ``portal'', which is a segment for 2D (or a triangle for 3D) that is inscribed to the Minkowski Difference. The illustrations from (a) to (c) show one MPR iteration, as the portal $(v_1, v_2)$ in (a) is updated to a new portal $(v_2, v_{\text{candidate}})$ in (c). }
\end{figure}

GJK/EPA \cite{gilbert1988gjk, van2001epa} uses an implicit, functional geometry representation called \textit{support function}. For a given convex set $A$, we can define its support function $\text{supp}_{A}(\cdot)$ as
\begin{align}
\begin{split}
\label{equ:suppdef}
    \text{supp}_{A}(d) = \text{argmin}_{~x \in A}~~\text{dot}(x, d)~ \subset A
\end{split}
\end{align}
\noindent where $\text{dot}(\cdot, \cdot)$ is dot product, and $d$ is a unit vector denoted as \textit{support direction}. Intuitively, the support function computes points in $A$ that are the farthest along the support direction. 
For some $d$, there can be more than one point that satisfies Equ.~(\ref{equ:suppdef}). Following existing works, we assume $\text{supp}_{A}(\cdot)$ produces one point for each $d$, which can be arbitrarily selected from points that satisfy Equ.~(\ref{equ:suppdef}).
%
The support function is closely related to the \textit{supporting hyperplane}~\cite{boyd2004convex} of a convex set. For a given direction $d$ and $v = \text{supp}_{A}(d)$, the plane $\{x|\text{dot}(d, x - v) = 0\}$ is a supporting plane for the convex shape $A$ at the point $v \in A$.

As shown in~\cite{gilbert1988gjk}, we can use $\text{supp}_{A_1}(\cdot)$ and $\text{supp}_{A_2}(\cdot)$ of two shapes $A_1,~A_2$ to construct the support function of their Minkowski Difference $\text{supp}_{\mathcal{D}_{1, 2}}(\cdot)$. In particular,
%
\begin{equation}
    \text{supp}_{\mathcal{D}_{1, 2}}(d) = \text{supp}_{A_{1}}(d) - \text{supp}_{A_{2}}(-d)
\end{equation}
\noindent 
In other words, we can get a point from $\text{supp}_{\mathcal{D}_{1, 2}}(d)$ using $\text{supp}_{A_1}(d)$ and $\text{supp}_{A_2}(-d)$.
Algorithms in this work will access the Minkowski Difference through its support function. 

\subsection{Origin Ray and Minkowski Portal Refinement}
\label{subsec:mprpre}
%
%
We investigate two overlapping shapes $A_1$ and $A_2$, thus their Minkowski Difference ${\mathcal{D}_{1, 2}}$ contains the origin. Given a direction $d_{\text{ray}}$ as the input, we define the \textit{origin ray} that starts from the origin $O$ and extends along $d_{\text{ray}}$ as
\begin{align}
\begin{split}
  \text{origin\_ray}(d_{\text{ray}}) =  \{x | x = t~d_{\text{ray}}, \text{~for~} t \geq 0 \}
\end{split}
\end{align}

The MPR~\cite{jacobs2008mpr} algorithm was originally proposed for binary collision checking. In this subsection, we present a modified MPR that computes the intersection point of the origin ray with the boundary of ${\mathcal{D}_{1, 2}}$:
\begin{align}
\begin{split}
  v_{\text{intersect}}(d_{\text{ray}}) = \text{origin\_ray}(d_{\text{ray}}) \cap \text{boundary}({\mathcal{D}_{1, 2}})
\end{split}
\end{align}

As illustrated in Fig.~\ref{fig:mpr_illustration} (a), MPR maintains and updates a \textit{portal}, which is a $n$-simplex inscribed to the Minkowski Difference ${\mathcal{D}_{1, 2}} \subset R^n$ that intersects with the origin ray. For 2D setup, the portal is a segment inscribed to ${\mathcal{D}_{1, 2}}$. For example, the segment $(v_1, v_2)$ in Fig.~\ref{fig:mpr_illustration} (a) is a portal that intersects the origin ray at $p$. In 3D setup, the portal is a triangle. The portal is initialized using a `find\_portal'' procedure~\cite{jacobs2008mpr}, as shown in Algorithm~\ref{alg:mpr}.

In each iteration, MPR computes the portal normal $d_\text{portal}$. Then, the support function $\text{supp}_{A}(\cdot)$ is invoked with $d_\text{portal}$ to get a point $v_{\text{candidate}}$. As the origin ray intersects with portal $(v_1, v_2)$, it must intersect with one of the segment $(v_1, v_{\text{candidate}})$ and $(v_2, v_{\text{candidate}})$. The intersecting one would be the new portal for the next iteration. This is the update\_portal in Algorithm~\ref{alg:mpr}. 
The iterations continue until $p$ is a good approximation of $v_{\text{intersect}}$, up to some tolerance $\Delta$ in the portal normal direction, as shown in Fig.~\ref{fig:mpr_illustration}.


\begin{algorithm}[t]
\caption{MPR for Origin Ray Intersection in Sec.~\ref{subsec:mprpre}}
\label{alg:mpr}
\begin{algorithmic}
\Require $\mathcal{D}_{1, 2}$ with support function $\text{supp}_{\mathcal{D}_{1, 2}}(\cdot)$
\Require ray direction $d_{\text{ray}}$
\Require tolerance $\Delta$
\State $\text{portal}_0 \gets \text{find\_portal}(\mathcal{D}_{1, 2})$
\While{$k = 0, 1, 2, ...$}
\State $d_{\text{portal}\_k} \gets \text{portal\_normal}(\text{portal}_k)$
\State $v_{\text{candidate}} \gets \text{supp}_{\mathcal{D}_{1, 2}}(\text{portal}_k)$
\State $p_k \gets \text{intersect}(\text{origin\_ray}(d_{\text{ray}}), d_{\text{portal}\_k})$
\State \Comment{A plane is defined by its normal and one point on it}
\State $q_k \gets \text{intersect}(\text{origin\_ray}(d_{\text{ray}}), \text{Plane}(d_{\text{portal}\_k}, v_{\text{candidate}}))$
\If{$|\text{dot}(q_k - p_k, \text{normalized}( d_{\text{portal}\_k}  ))| \leq \Delta$}
    \State \textbf{return} ($p_k$, $v_{\text{candidate}}$, $d_{\text{portal}\_k}$)
\EndIf
\State $\text{portal}_{k+1} \gets \text{update\_portal}(\text{portal}_{k},~v_{\text{candidate}})$
\EndWhile
\end{algorithmic}
\end{algorithm}

\section{Penetration Depth Estimation Algorithm}
\label{sec:algorithm}

As shown in Sec.~\ref{sec:preliminary}, PD estimation can be expressed as
\begin{align}
\begin{split}
\label{equ:actualopt}
     &\text{min}_{~v \in R^n}~~~||v||_2 \\
     &\text{subject to:}~~v \in \text{boundary}(\mathcal{D})
\end{split}
\end{align}
\noindent where $\mathcal{D}$ is the Minkowski Difference that contains the origin. The subscript of ${\mathcal{D}_{1, 2}}$ is dropped for notational simplicity. 
%
As shown in Sec.~\ref{subsec:mkdiffpre}, we can only access the convex set $\mathcal{D}$ through its support function $\text{supp}_{\mathcal{D}}(\cdot)$. This implicit geometric representation implies:
\begin{enumerate}
    \item For a given point $v \in \text{boundary}(\mathcal{D})$, it is hard to compute a supporting hyperplane of $\mathcal{D}$ (or equivalently the hyperplane \textit{normal}) that passes through $v$. This operation is required for many optimization procedures, such as the SQP in Sec.~\ref{subsec:alg_1}.
    \item The support function $\text{supp}_{\mathcal{D}}(\cdot)$ is actually an ``inverse'' of the operation in 1): for a given normal direction $d$, we can compute a point $v = \text{supp}_{\mathcal{D}}(d)$. This point, combined with the normal $d$, is able to define a supporting hyperplane for $\mathcal{D}$.
\end{enumerate}
\noindent In this section we will address this challenge and propose an optimization-based PD estimation algorithm.

To make a clear presentation, we first introduce two ``conceptual'' algorithms in Sec.~\ref{subsec:alg_1} and~\ref{subsec:alg_2}. The first one highlights the SQP procedure assuming we can compute supporting hyperplanes for $v \in \text{boundary}(\mathcal{D})$. As mentioned above, this assumption is not true for our problem. 
To address it, we propose a novel instantiation of SQP that utilizes a modified MPR algorithm as a subroutine. This is the second ``conceptual'' algorithm in Sec.~\ref{subsec:alg_2}.
Finally, we present the actual PD algorithm with a shortcut mechanism, which reduces the computation and leads to substantial speed-up.

\subsection{Formulation as a Difference-of-Convex Problem}
\label{subsec:alg_1}

In this subsection, we investigate the following optimization problem, which is very similar to Problem~(\ref{equ:actualopt}):
\begin{align}
\begin{split}
\label{equ:concaveopt}
     &\text{min}_{~v \in R^n}~~~||v||_2 \\
     &\text{subject to:}~~v \in \text{boundary}(\mathcal{E})
\end{split}
\end{align}
\noindent where $\mathcal{E}$ is also a closed convex set in $R^n$ that contains the origin. Different from Problem~(\ref{equ:actualopt}), we assume an explicit representation of $\mathcal{E}$ is available, such that we can compute a supporting hyperplane for each point $v \in \text{boundary}(\mathcal{E})$. Since this assumption does not hold for the Minkowski Difference $\mathcal{D}$, the PD algorithm presented in this subsection is ``conceptual'' and will be instantiated in Sec.~\ref{subsec:alg_2}.

As a concrete example, the convex shape $\mathcal{E}$ might be represented as the level-set of a continuous convex function $g(v)$. In other words, $\mathcal{E} = \{v | g(v) \leq 0 \}$ and the boundary is $\{v | g(v) = 0 \}$. For a point $v \in \text{boundary}(\mathcal{E})$, we can compute a supporting hyperplane whose normal is $\nabla g(v)$, where $\nabla$ is the subgradient operator.

The domain for the decision variable in Problem~(\ref{equ:concaveopt}) only includes the boundary of $\mathcal{E}$. As the convex set $\mathcal{E}$ contains the origin, we can enlarge the input domain to everywhere except the inner of $\mathcal{E}$ and rewrite the optimization as
\begin{align}
\begin{split}
\label{equ:concaveoptexternal}
     &\text{min}_{~v \in R^n}~~~||v||_2 \\
     &\text{subject to:}~~v \in (R^n \setminus \text{inner}(\mathcal{E}))
\end{split}
\end{align}
\noindent where $R^n \setminus \text{inner}(\mathcal{E})$ is the complement of $\text{inner}(\mathcal{E})$.
Problem~(\ref{equ:concaveoptexternal}) can be solved using the SQP~\cite{boyd2004convex} algorithm. Moreover, Problem~(\ref{equ:concaveoptexternal}) is a Difference-of-Convex problem. Thus, additional properties can be used to simplify the SQP:
\begin{itemize}
    \item SQP usually requires a \textit{trust region}. This trust region can be infinity for Difference-of-Convex problems~\cite{boyd2004convex}.
    \item The linearization of the constraint $v \in (R^n \setminus \text{inner}(\mathcal{E}))$ is a halfspace outside the supporting hyperplane. Thus, the inner QP step of SQP can be implemented by projecting the origin onto the supporting hyperplane.
\end{itemize}

\noindent The SQP algorithm is presented in Algorithm~\ref{alg:sqp1}, which alternates between: 1) compute the supporting hyperplane; 2) project the origin onto the hyperplane; and 3) compute the intersection of the projection line with the boundary. Detailed derivation is provided in the Supplemental Material.


\begin{algorithm}[t]
\caption{SQP for Problem~\ref{equ:concaveoptexternal}}
\label{alg:sqp1}
\begin{algorithmic}
\Require $\mathcal{E}$ that supports compute\_supporting\_hyperplane($\cdot$)
\Require $v_{\text{init}} \in \text{boundary}(\mathcal{E})$
\State $v_0 \gets v_{\text{init}}$
\While{$k = 0, 1, 2, ...$}
\State $n_{\text{plane}\_k} \gets \text{compute\_supporting\_hyperplane}(\mathcal{E},~v_k)$
\State \Comment{Plane defined by normal $n_{\text{plane}\_k}$ and a point $v_k$ on it}
\State $z_k \gets \text{project\_origin\_to\_plane}(\text{Plane}(v_k,~n_{\text{plane}\_k}))$
\State $v_{k+1} = \text{boundary\_intersection}(O\_\text{to}\_z_k,~\text{boundary}(\mathcal{E}))$
\EndWhile
\State \textbf{return} $v_k$
\end{algorithmic}
\end{algorithm}


Algorithm~\ref{alg:sqp1} cannot be directly applied to Minkowski Difference $\mathcal{D}$, as it requires explicit geometric representation to compute a supporting hyperplane for boundary points. This challenge will be addressed in Sec.~\ref{subsec:alg_2}.

\subsection{Instantiation using MPR as a Subroutine}
\label{subsec:alg_2}

In this subsection, we adopt the SQP procedure in Algorithm~\ref{alg:sqp1} to the Minkowski Difference $\mathcal{D}$, which is represented by its support function $\text{supp}_{\mathcal{D}}(\cdot)$. For each point $v \in \text{boundary}(\mathcal{D})$, we can compute its direction $d_v = \text{normalized}(v)$. 
We propose to use $d_v$ as the decision variable and represent $v$ in Problem~(\ref{equ:concaveoptexternal}) implicitly by $d_v$, based on the following observations:
\begin{enumerate}
    \item The point $v \in \text{boundary}(\mathcal{D})$ can be estimated from $d_v$ (with accuracy up on some tolerance $\Delta$), using the MPR algorithm in Sec.~\ref{subsec:mprpre} as a subroutine.
    \item More importantly, the supporting hyperplane normal at $v$ can be approximated by the portal normal $d_{\text{portal}}$ upon the convergence of the MPR algorithm. It is emphasized that usually $d_{\text{portal}} \neq d_v$.
\end{enumerate}
\noindent With these observations, we propose Algorithm~\ref{alg:sqp2} for the Minkowski Difference $\mathcal{D}$, which is transformed from Algorithm~\ref{alg:sqp1}. As this algorithm will be superseded in Sec.~\ref{subsec:alg_3}, we only provide an intuitive analysis in this subsection.

\begin{figure}[t]
\centering
\includegraphics[width=0.48\textwidth]{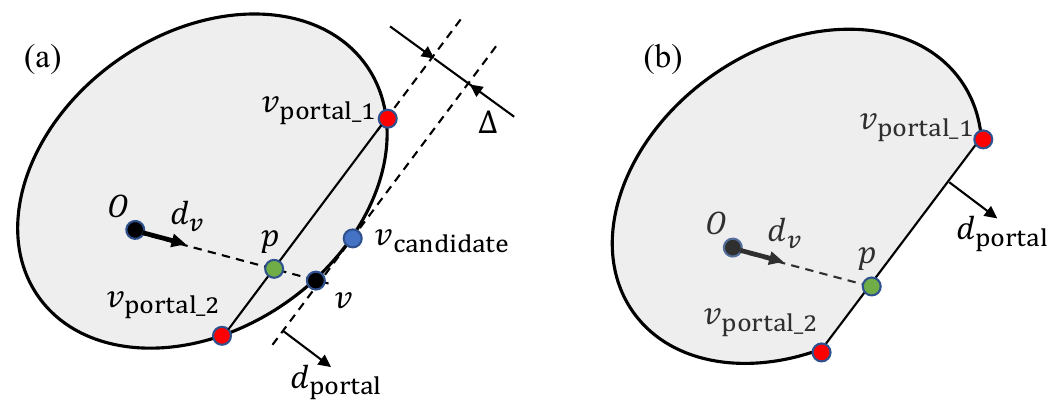}
\caption{\label{fig:approximation} Approximation scheme used in Sec.~\ref{subsec:alg_2} and Algorithm~\ref{alg:sqp2}. (a) shows the configuration of the portal and supporting hyperplane upon the convergence (with tolerance $\Delta$) of MPR, when it is used as a subroutine in Algorithm~\ref{alg:sqp2}. Intuitively, the approximation in Sec.~\ref{subsec:alg_2} corresponds to removing all points ``outside'' the final portal, as shown in (b). A detailed explanation is in Sec.~\ref{subsec:alg_2}. }
\end{figure}

A critical property of Algorithm~\ref{alg:sqp2} is approximating a supporting hyperplane normal by $d_{\text{portal}\_k}$, which is further illustrated in Fig.~\ref{fig:approximation}. Upon the convergence of MPR (Algorithm~\ref{alg:mpr}) subroutine with tolerance $\Delta$, the ground-truth intersecting point $v$ for direction $d_v$ must lie in the between of the portal (segment $(v_{\text{portal}\_1}, v_{\text{portal}\_2})$ in Fig.~\ref{fig:approximation}) and the supporting hyperplane at $v_{\text{candidate}}$. The portal and supporting hyperplane are parallel to each other (they share the same normal $d_{\text{portal}}$), and the distance between these two planes is no more than $\Delta$ as the MPR terminates in this iteration. Algorithm~\ref{alg:mpr} uses $p$ to approximate $v$ and uses $d_{\text{portal}}$ to approximate a supporting hyperplane normal at $v$. 
This approximation is equivalent to removing points from $\mathcal{D}$ that are separated from the origin by the portal, as illustrated in Fig.~\ref{fig:approximation} (b).
This is because $p$ is the exact intersecting point for the shape in Fig.~\ref{fig:approximation} (b), while $d_{\text{portal}}$ is the exact supporting hyperplane at that intersecting point. The removed part is a convex region whose ``thickness'' along $d_{\text{portal}}$ is no more than $\Delta$. Thus, the approximation becomes more accurate as the tolerance $\Delta$ is set smaller.

Algorithm~\ref{alg:sqp2} might work in practice. However, there can be many iterations in the MPR subroutine, which is computationally expensive. This issue will be addressed in Sec.~\ref{subsec:alg_3}.
\begin{algorithm}[t]
\caption{SQP using Support Function}
\label{alg:sqp2}
\begin{algorithmic}
\Require support function $\text{supp}_{\mathcal{D}}(\cdot)$
\Require initial direction $d_{\text{init}}$
\Require tolerance $\Delta$ for MPR subprocedure
\State $d_0 \gets d_{\text{init}}$
\While{$k = 0, 1, 2, ...$}
\State $\text{mpr\_output} \gets \text{mpr}(\text{supp}_{\mathcal{D}},~d_k,~\Delta)$ \Comment{Algorithm~\ref{alg:mpr}}
\State $(p_k,~v_{\text{candidate}},~d_{\text{portal}\_k}) \gets \text{mpr\_output}$
\State $\text{support\_plane} \gets \text{Plane}(v_{\text{candidate}},~d_{\text{portal}\_k})$
\State $z_k \gets \text{project\_origin\_to\_plane}(\text{support\_plane})$
\If{ should\_terminate($p_k,~z_k,~d_{\text{portal}\_k}$) }
    \State \textbf{return} $\text{intersect}(\text{origin\_ray}(d_k), \text{support\_plane})$
\EndIf
\State $d_{k+1} \gets d_{\text{portal}\_k}$
\EndWhile
\end{algorithmic}
\end{algorithm}

\subsection{Penetration Depth Estimation with Shortcut}
\label{subsec:alg_3}

In existing software stacks~\cite{pan2012fcl, coumans2016pybullet}, MPR is typically used with a rather small tolerance for accurate binary collision checking. However, the primary functionality for the MPR subroutine in Algorithm~\ref{alg:sqp2} is to produce a search direction $d_{k+1}$ via the supporting hyperplane normal $d_{\text{portal}}$.
Thus, we propose to reduce the MPR iterations by an early termination mechanism (shortcut) once a ``good'' $d_{\text{portal}}$ is discovered.

\begin{figure}[t]
\centering
\includegraphics[width=0.23\textwidth]{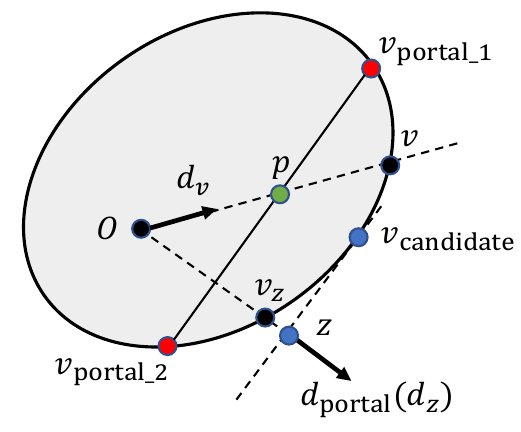}
\caption{\label{fig:shortcut} An illustration of the shortcut mechanism in Sec.~\ref{subsec:alg_3}. The penetration depth along $d_v$ is lower bounded by $|p|$, while the penetration depth along $d_{\text{portal}}(d_z)$ is upper bounded by $z$. We proposed to switch to $d_z$ as the new search direction when $|p| 
\geq |z|$, as explained in Sec.~\ref{subsec:alg_3}. }
\end{figure}

Fig.~\ref{fig:shortcut} shows the configuration of the portal and supporting hyperplane in the MPR subroutine. The distance between the portal and the supporting hyperplane might be (much) larger than the tolerance $\Delta$, as it is not necessarily the final iteration. Let the point $p$ be the intersection of $\text{origin\_ray}(d_v)$ with the portal. MPR subroutine uses $p$ as the current estimation of $v$ (the intersection with the boundary of $\mathcal{D}$), and $|p| \leq |v|$. In other words, if we continue the MPR subroutine with direction $d_v$, the PD we would find is \textit{lower} bounded by $|p|$.

Besides, we can project origin $O$ to the supporting hyperplane to get $z$, as shown in Fig~\ref{fig:shortcut}. Let $v_z$ be the intersection of segment $(O, z)$ with the boundary of $\mathcal{D}$. Obviously $|v_z| \leq |z|$. If we switch to a new search direction $d_z = \text{normalized}(z)$, the PD we would find is \textit{upper} bounded by $|z|$.

From the analysis above, in each MPR subroutine iteration, we have: 1) a lower bound of PD $|p|$ along the current search direction $d_v$; and 2) a new search direction $d_z$ candidate with an upper bound on PD $|z|$ along it. We propose to switch to the new search direction $d_z$ once the upper bound $z$ along $d_z$ is smaller than the lower bound along the original direction $d_v$, as shown in Fig.~\ref{fig:shortcut}.

With the analysis above, the overall PD estimation algorithm is summarized in Algorithm~\ref{alg:sqp3}, which uses a subroutine in Algorithm~\ref{alg:mpr_shortcut}. The SQP procedure is almost identical to Algorithm~\ref{alg:sqp2}, except that the MPR subroutine is changed to Algorithm~\ref{alg:mpr_shortcut}. The MPR subroutine would be terminated early with the abovementioned condition, as shown in Algorithm~\ref{alg:mpr_shortcut}. Despite the new shortcut mechanism, this SQP is guaranteed to converge to a locally optimal solution. Please refer to the Supplemental Material for a detailed analysis.

\begin{algorithm}[t]
\caption{Proposed SQP Algorithm for PD Estimation}
\label{alg:sqp3}
\begin{algorithmic}
\Require support function $\text{supp}_{\mathcal{D}}(\cdot)$
\Require initial direction $d_{\text{init}}$
\State Replaces the invoked MPR subroutine in Algorithm~\ref{alg:sqp2} by mpr\_shortcut in Algorithm~\ref{alg:mpr_shortcut}. The tolerance $\Delta$ in Algorithm~\ref{alg:sqp2} is no longer necessary.
\end{algorithmic}
\end{algorithm}

\subsection{Implementation Details: Initialization}
\label{subsec:alg_impl}

The Algorithm~\ref{alg:sqp3} requires an initial direction $d_{\text{init}}$. This is a guess of the smallest displacement direction that could move shape $A_1$ away from colliding with shape $A_2$. For applications that exhibit high spatial or temporal coherence, there might be a good application-specific guess of the minimum penetration direction. For instance, in dynamic simulation~\cite{trinkle1997dynamic, mirtich1998rigid, todorov2014convex} colliding object pairs tend to have very similar penetration direction and depth in consecutive simulation steps. Similarly, in optimization-based motion planning~\cite{toussaint2018differentiable, schulman2013finding, kappler2018real} we might use the penetration direction from the previous optimization iteration as $d_{\text{init}}$, especially for planning algorithms with trust-region mechanisms to ensure a small difference between consecutive iterations.
If application-specific prior about penetration direction is unavailable, one alternative method is to use the centroid difference between two shapes as $d_{\text{init}}$ as suggested by~\cite{kim2004incremental}. 

\begin{algorithm}[t]
\caption{MPR Subroutine with Shortcut in Sec.~\ref{subsec:alg_3}}
\label{alg:mpr_shortcut}
\begin{algorithmic}
\Require $\mathcal{D}_{1, 2}$ with support function $\text{supp}_{\mathcal{D}_{1, 2}}(\cdot)$
\Require ray direction $d_{\text{ray}}$
\State $\text{portal}_0 \gets \text{find\_portal}(\mathcal{D}_{1, 2})$
\While{$k = 0, 1, 2, ...$}
\State $d_{\text{portal}\_k} \gets \text{portal\_normal}(\text{portal}_k)$
\State $v_{\text{candidate}} \gets \text{supp}_{\mathcal{D}_{1, 2}}(\text{portal}_k)$
\State $p_k \gets \text{intersect}(\text{origin\_ray}(d_{\text{ray}}), \text{portal}_{k})$
\State \Comment{A plane is defined by its normal and one point on it}
\State $z_k \gets \text{project\_origin\_to\_plane}(\text{Plane}(v_{\text{candidate}},~d_{\text{portal}\_k}))$
\If{ $|z_k| \leq |p_k|$ } \Comment{Shortcut in Sec.~\ref{subsec:alg_3}}
    \State \textbf{return} $p_k$, $v_{\text{candidate}}$, $d_{\text{portal}\_k}$
\EndIf
\State $\text{portal}_{k+1} \gets \text{update\_portal}(\text{portal}_{k},~v_{\text{candidate}})$
\EndWhile
\end{algorithmic}
\end{algorithm}

\section{Results}
\label{sec:results}

In this section, we experimentally investigate the robustness, accuracy and efficiency of the proposed method. We first compare our method with EPA~\cite{van2001epa} and DEEP~\cite{kim2004incremental}. Then, an ablation study is conducted to highlight the shortcut mechanism introduced in Sec.~\ref{subsec:alg_3}.
Parameters are tuned such that all algorithms have roughly the same accuracy. 
The source code and supplemental document are available on \href{https://github.com/weigao95/mind-fcl}{this link}.

\subsection{Comparison with EPA}
\label{subsec:epa_compare}

Our method is compared with the EPA~\cite{van2001epa} algorithm implemented in libccd~\cite{libccd}, a popular library used in various robotics applications~\cite{pan2012fcl, todorov2014convex}. The evaluation is performed on two sets of shapes. The first one consists of simple primitives, and the second set is convex polyhedra with different numbers of vertices.

\vspace{1mm}
\noindent \textbf{Primitive Shapes. }
We consider the penetration depth between three types of shape pairs:

\begin{itemize}
    \item Sphere collides with sphere (sphere vs. sphere)
    \item Capsule collides with capsule (capsule vs. capsule)
    \item Sphere collides with capsule (sphere vs. capsule)
\end{itemize}

\noindent For these simple primitive shapes, we use hand-written PD algorithms as the ground truth. The shapes have characteristic dimensions (such as sphere diameter) of 1 meter in this experiment. All of the statistical results are the averaged value of 10000 independent runs with randomly generated poses for the shape pair. Please refer to supplemental materials for detailed setup.

Our method requires an initial penetration direction guess. For the purpose of benchmarking, we use the following method to obtain the initial guess. Given the ground truth penetration direction from the hand-written algorithm, we apply a rotation of $n$ degrees with respect to a random axis that is perpendicular to the ground-truth direction. Then, the rotated direction is used as the initial guess.

Table.~\ref{table:table_error_1} and~\ref{table:table_error_2} shows the accuracy of our method under different rotation perturbation angles $n$, where $n$ ranges from $5^{\circ}$ to $45^{\circ}$. Table.~\ref{table:table_error_1} presents the distance error while Table.~\ref{table:table_error_2} shows the directional error. We report the average deviation of 10000 independent runs. Both EPA and our method can accurately estimate the penetration. The proposed algorithm is rather accurate despite the large initialization error. 

Table.~\ref{table:table_speed_1} summarizes the performance of our method. Our method achieves 20x-30x speed up compared to EPA. Moreover, the performance improvement is consistent for different shape types and deviations of initial guess direction.

\begin{table}[t]
\centering
\begin{tabular}{ |m{1.3cm}|m{0.9cm}|m{0.9cm}|m{0.9cm}|m{0.9cm}|}
\hline
& Ours ($5^{\circ}$) & Ours ($25^{\circ}$) & Ours ($45^{\circ}$) & EPA \\
\hline
Sphere vs. Sphere & 0.909& 1.018 & 1.03 & 1.58 \\
\hline
Capsule vs. Capsule & 1.092 & 1.12 & 1.12 & 0.39 \\
\hline
Sphere vs. Capsule & 1.255 & 1.31 & 1.30 & 1.72 \\
\hline
\end{tabular}
\caption{ The penetration depth error in micrometers ($10^{-6}$ meter) of various algorithms compared with the ground truth. The result errors are the average of 10000 independent runs. Our method is evaluated with different angular deviations (of initial penetration direction) at $5^{\circ}$, $25^{\circ}$, and $45^{\circ}$. Geometries have characteristic dimensions (such as the diameter of the sphere) of 1 meter. }
\label{table:table_error_1}
\end{table}

\begin{table}[t]
\centering
\begin{tabular}{ |m{1.3cm}|m{0.9cm}|m{0.9cm}|m{0.9cm}|m{0.9cm}|}
\hline
& Ours ($5^{\circ}$) & Ours ($25^{\circ}$) & Ours ($45^{\circ}$) & EPA \\
\hline
Sphere vs. Sphere & 1.82 & 1.38 & 1.93 & 8.84 \\
\hline
Capsule vs. Capsule & 3.19 & 3.42 & 3.23 & 10.39 \\
\hline
Sphere vs. Capsule & 2.25 & 2.31 & 2.30 & 10.72 \\
\hline
\end{tabular}
\caption{ The penetration direction error in milliradian ($10^{-3}$ radian) of various algorithms compared with the ground truth. The result errors are the average of 10000 independent runs. Our method is evaluated with different angular deviations (of initial penetration direction) at $5^{\circ}$, $25^{\circ}$, and $45^{\circ}$. }
\label{table:table_error_2}
\end{table}

\begin{table}
\centering
\begin{tabular}{ |m{1.3cm}|m{0.7cm}|m{0.7cm}|m{0.7cm}|m{0.8cm}|}
\hline
& Ours ($5^{\circ}$) & Ours ($25^{\circ}$) & Ours ($45^{\circ}$) & EPA \\
\hline
Sphere vs. Sphere & 1.9 & 2.1 & 2.3 & 85.1 \\
\hline
Capsule vs. Capsule & 1.6 & 1.9 & 2.0 & 39.5 \\
\hline
Sphere vs. Capsule & 2.0 & 2.2 & 2.4 & 61.5 \\
\hline
\end{tabular}
\caption{ Performance of the proposed method under different angular deviations of the initial direction guess. Results are the averaged time in microseconds from 10000 independent runs. The proposed method is 20x-30x times faster than EPA.}
\label{table:table_speed_1}
\end{table}

\begin{figure}[t]
\centering
\includegraphics[width=0.3\textwidth]{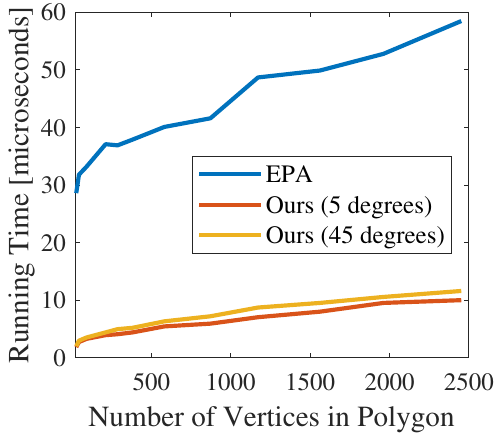}
\caption{\label{fig:speed_nv} Performance comparison between our method and EPA~\cite{van2001epa} under different number of vertices in convex polyhedron. Results are the averaged time in microseconds computed from 10000 independent runs. Our method is evaluated with different angular deviations (of initial penetration direction) at $5^{\circ}$ and $45^{\circ}$. The proposed method achieves 6-10x speedup compared to the EPA baseline. }
\end{figure}

\vspace{1mm}
\noindent \textbf{Convex polyhedra. }
The second experiment compares our method with EPA on convex polyhedra. We use the similar sphere-capsule setup, but replace the sphere with its convex polyhedral approximation at different resolutions. Higher resolution with more vertices and faces improves the approximation accuracy at the cost of computational complexity.

Fig.~\ref{fig:speed_nv} shows the performance of the proposed algorithm with respect to the number of vertices in the convex polyhedra. 
The proposed method consistently outperforms EPA baselines with about 6-10x speedup.

\subsection{Comparison with DEEP~\cite{kim2004incremental}}

The proposed method is compared with~\cite{kim2004incremental}, which computes penetration distance incrementally by walking on the surface of Minkowski Difference. We implement the baseline algorithm in C++, as well as the internal convex polygon intersection algorithm~\cite{o1982new}. We use the suggested initialization scheme: given an initial guess of penetration direction, compute the vertex hub pair by taking the extremal vertex of each shape along that direction. As \cite{kim2004incremental} can only handle convex polyhedra, we use a sphere-sphere setup similar to Sec.~\ref{subsec:epa_compare} but replace the sphere with its convex polyhedral approximation.

\begin{figure}[t]
\centering
\includegraphics[width=0.3\textwidth]{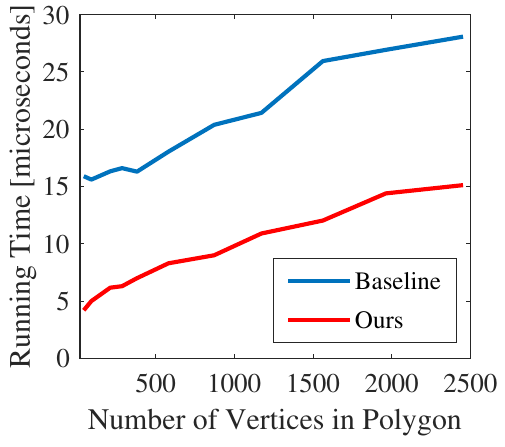}
\caption{\label{fig:speed_nv2} Performance comparison between our method and DEEP~\cite{kim2004incremental} under different number of vertices in convex polyhedron. Results are the averaged time in microseconds. Both methods are evaluated with angular deviation (of initial penetration direction) at $5^{\circ}$. The proposed method is about 2x faster compared to the baseline. }
\end{figure}

Fig.~\ref{fig:speed_nv2} shows the performance comparison between the proposed algorithm and the baseline. The initial direction deviation is fixed as $5^{\circ}$ and the sphere shape is discretized at different resolutions. The proposed method outperforms the baseline~\cite{kim2004incremental} with about 2x speedup.

\subsection{Ablation Study on Shortcut Mechanism}

In Sec.~\ref{subsec:alg_3}, a shortcut mechanism is introduced to improve the performance by reducing the number of MPR iterations. To highlight its benefit, we conduct an ablation study by removing the shortcut mechanism of the proposed algorithm. We use the same convex polyhedra setup as Sec.~\ref{subsec:epa_compare}. The initial direction deviation is fixed as $45^{\circ}$.

\begin{figure}[t]
\centering
\includegraphics[width=0.485\textwidth]{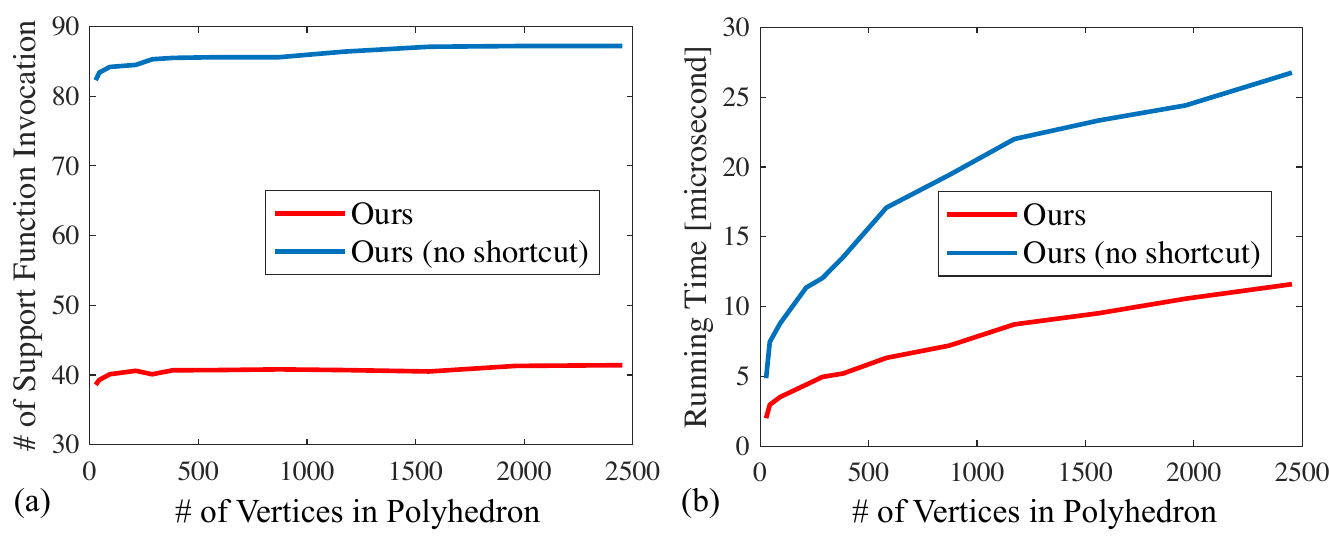}
\caption{\label{fig:no_shortcut} An ablation study is conducted to highlight the benefit of the proposed shortcut in Sec.~\ref{subsec:alg_3}. In the figure, (a) shows the performance measured by the number of support function invocations, while (b) shows the performance in microseconds. The proposed shortcut mechanism can halve the required support function invocations and lead to about 2-3x speedup. }
\end{figure}

The result is shown in Fig.~\ref{fig:no_shortcut}. The number of support function invocations and running time are used to characterize the performance. From the figure, the proposed shortcut mechanism can halve the required support function invocations and lead to about 2-3x speedup.


\section{Conclusion}
\label{sec:conclusion}

This paper presents a novel algorithm to estimate the PD between convex shapes. To achieve this, we formulate the PD estimation as a Difference-of-Convex optimization. Then, we propose a novel instantiation of SQP using a modified MPR subroutine that solves the optimization-based PD estimation. We further present a shortcut mechanism that significantly reduces the computation. Through various experiments, we show that the proposed method achieves a 5-30x speedup over the well-known EPA algorithm at comparable accuracy.

\section{Acknowledgments}

This work was conducted during the author's employment at Mech-Mind Robotics. The views expressed in this paper are those of the authors themselves and are not endorsed by the supporting agencies.

{\small
\bibliographystyle{abbrv}
\bibliography{paper.bib}
}

\end{document}